\documentclass[conference]{IEEEtran}
\IEEEoverridecommandlockouts
\usepackage{cite}
\usepackage{amsmath,amssymb,amsfonts}
\usepackage{algorithmic}
\usepackage{graphicx}
\usepackage{textcomp}
\usepackage{xcolor}
\usepackage{hyperref}

\def\BibTeX{{\rm B\kern-.05em{\sc i\kern-.025em b}\kern-.08em
    T\kern-.1667em\lower.7ex\hbox{E}\kern-.125emX}}
\begin{document}

\title{Multi-Sample Dropout for Accelerated Training\\
 and Better Generalization}

\author{\IEEEauthorblockN{Hiroshi Inoue}
\IEEEauthorblockA{\textit{IBM Research - Tokyo} 
Tokyo, Japan \\
inouehrs@jp.ibm.com}
}

\maketitle

\begin{abstract}
Dropout is a simple but efficient regularization technique for achieving better generalization of deep neural networks (DNNs); hence it is widely used in tasks based on DNNs. During training, dropout randomly discards a portion of the neurons to avoid overfitting. This paper presents an enhanced dropout technique, which we call multi-sample dropout, for both accelerating training and improving generalization over the original dropout. The original dropout creates a randomly selected subset (called a dropout sample) from the input in each training iteration while the multi-sample dropout creates multiple dropout samples. The loss is calculated for each sample, and then the sample losses are averaged to obtain the final loss. This technique can be easily implemented by duplicating a part of the network after the dropout layer while sharing the weights among the duplicated fully connected layers. Experimental results using image classification tasks including ImageNet, CIFAR-10, and CIFAR-100 showed that multi-sample dropout accelerates training. Moreover, the networks trained using multi-sample dropout achieved lower error rates compared to networks trained with the original dropout. %The additional computation cost due to the duplicated operations is not significant for deep convolutional networks because most of the computation time is consumed in the convolution layers before the dropout layer, which are not duplicated.
\end{abstract}


\section{Introduction}

\begin{figure*}
  \begin{center}
    \includegraphics[width=12cm]{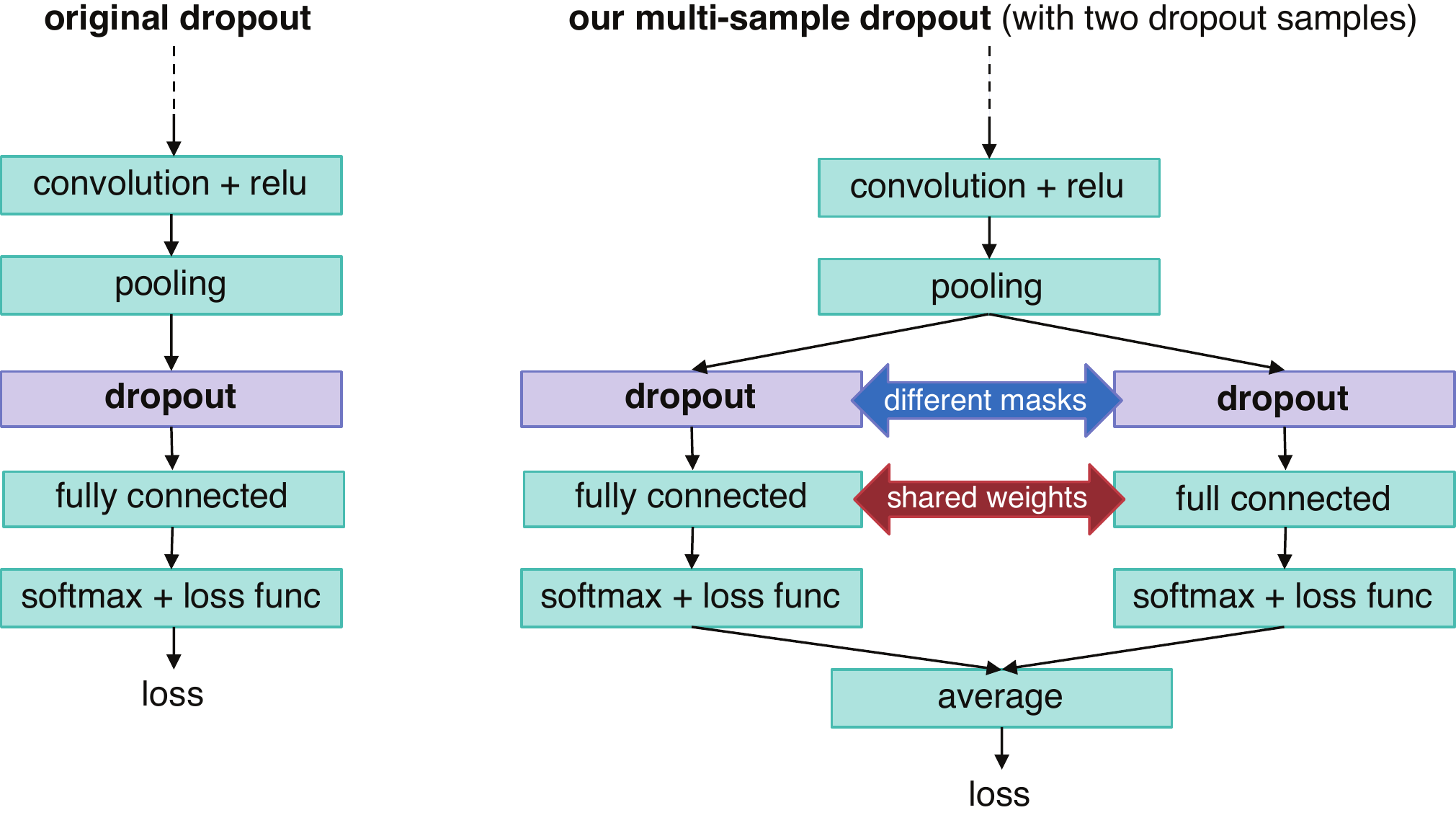}
    \caption{Overview of original dropout and our multi-sample dropout. }
  \end{center}
  \label{overview}
\end{figure*}

Dropout \cite{Hinton12} is one of the key regularization techniques for improving the generalization of deep neural networks (DNNs). Because of its simplicity and efficiency, the original dropout and various similar techniques are widely used to train neural networks for various tasks. The use of dropout prevents the trained network from overfitting to the training data by randomly discarding (i.e., "dropping") 50\% of the neurons at each training iteration. As a result, the neurons cannot depend on each other, and the trained network achieves better generalization. During inference, neurons are not discarded, so all information is preserved; instead, each outgoing value is multiplied by 0.5 to make the average value consistent with the training time. The network used for inference can be viewed as an ensemble of many sub-networks randomly created during training. The success of dropout inspired the development of many techniques using various ways for selecting information to discard. For example, DropConnect \cite{Wan13} discards a portion of the connections between neurons randomly selected during training instead of randomly discarding neurons. 

This paper reports {\it multi-sample dropout}, a dropout technique extended in a different way especially for deep convolutional neural networks (CNNs). The original dropout creates a randomly selected subset (a {\it dropout sample}) from the input during training. The proposed multi-sample dropout creates multiple dropout samples. The loss is calculated for each sample, and then the sample losses are averaged to obtain the final loss used for back propagation. By calculating losses for $M$ dropout samples and ensembling them, network parameters are updated to achieve smaller loss with any of these samples. This is similar to performing $M$ training repetitions for each input image in the same minibatch. Therefore, it significantly reduces the number of iterations needed for training. With our multi-sample dropout, we do not discard neurons during the inference, as with the original dropout. %Experiments demonstrated that it achieves smaller losses and errors for both the training set and validation set on image classification tasks. % partially because of its inherent ensemble mechanism. 

We observed that multi-sample dropout also improved accuracy of the trained network with increasing number of the dropout samples. Noh et al. \cite{Noh17} showed that creating multiple noise samples by noise injection, such as dropout, during the training of deep networks makes stochastic gradient descent (SGD) optimizers have a tighter lower bound in the marginal likelihood over noise. Our multi-sample dropout is an easy and effective way to exploit the benefits of using multiple noise samples without adding huge computation overhead.

%As far as the authors know, no other dropout technique uses a similar approach to accelerate training. 
In CNNs, dropout is typically applied to layers near the end of the network. VGG16 \cite{Simonyan2014}, for example, uses dropout for 2 fully connected layers following 13 convolution layers. Because the execution time for the fully connected layers is much shorter than that for the convolution layers, duplicating the fully connected layers for each of multiple dropout samples does not significantly increase the total execution time per iteration. Experiments using the ImageNet, CIFAR-10, and CIFAR-100 datasets showed that, with an increasing number of dropout samples created at each iteration, the improvements obtained (reduced number of iterations needed for training) became more significant at the expense of a longer execution time per iteration. % with up to 64 dropout samples.
Consideration of the reduced number of iterations along with the increased time per iteration revealed that the total training time was the shortest with a moderate number of dropout samples, such as 8. %In the training of a small CNN using multi-sample dropout with eight dropout samples and CIFAR-10 data, for example, the number of iterations until convergence was reduced half that with the original dropout while the execution time per iteration was increased by 19.0\%. The validation error was reduced from 8.1\% to 7.4\%., Similar improvements were observed for ImageNet using VGG16 as the network. 

Multi-sample dropout can be easily implemented on various existing deep learning frameworks without adding a new operator by duplicating a part of the network after the dropout layer while sharing the weights among the fully connected layers duplicated for each dropout sample. 

The main contribution of this paper is multi-sample dropout, a new regularization technique for accelerating the training of deep neural networks compared to the original dropout. Evaluation of multi-sample dropout on image classification tasks % using the ImageNet, CIFAR-10, CIFAR-100, and SVHN datasets 
demonstrated that it increases accuracy for both the training and validation sets as well as accelerating the training. %The multi-sampling technique should be also applicable to other regularization techniques based on random omission, such as DropConnect. 

\section{Multi-Sample Dropout}

\subsection{Overview}

This section describes the multi-sample dropout technique. The basic idea is quite simple: create multiple dropout samples instead of only one. Figure 1 depicts an easy way to implement multi-sample dropout (with two dropout samples) using an existing deep learning framework with only common operators. The dropout layer and several layers after the dropout are duplicated for each dropout sample; in the figure, the "dropout," "fully connected," and "softmax + loss func" layers are duplicated. Different masks are used for each dropout sample in the dropout layer so that a different subset of neurons is used for each dropout sample. In contrast, the parameters (i.e., connection weights) are shared between the duplicated fully connected layers. The loss is computed for each dropout sample using the same loss function, e.g., cross entropy, and the final loss value is obtained by averaging the loss values for all dropout samples. This final loss value is used as the objective function for optimization during training. We select the class label as the prediction based on the average of outputs from the last fully connected layer. Although a configuration with two dropout samples is shown in Figure 1, multi-sample dropout can be configured to use any number of dropout samples. The original dropout can be seen as a special case of multi-sample dropout where the number of samples is set to one.
%Increasing the number of samples increases accuracy at the expense of computation time per iteration and memory consumption. 

During inference, neurons are not discarded as is done in the original dropout. The loss can be calculated using only one dropout sample because the dropout samples become identical at the inference time if we do not drop any neurons at the dropout layer. Hence, we always use only one dropout sample at inference regardless of the training method. %, enabling the network to be pruned to eliminate redundant computations. %Note that, using all the dropout samples at the inference time does not badly affect the prediction performance, it just slightly increases the inference-time computation costs.

Compared to Importance Weighted Stochastic Gradient Descent (IWSGD) \cite{Noh17}, which also makes multiple samples by dropout, we only duplicate operations in a part of the forward pass after the dropout while IWSGD duplicate operations in the entire backward pass as well as the forward pass. Hence our multi-sample dropout is much more light weight in terms of computation costs. Especially when dropout is applied to a layer near the end of the network, the additional execution time due to the duplicated operations in multi-sample dropout is not significant; this characteristic makes multi-sample dropout more suitable for deep CNNs. We can apply multi-sample dropout for shallow networks, such as the multilayer perceptron. We observed that multi-sample dropout reduces the number of iterations for training even for shallow networks, but the costs of the increased execution time per iteration surpassed the benefits; due to the increase in the computation time per iteration, multi-sample dropout actually degraded the training speed in terms of the computation time.

If the network includes multiple dropout layers, we can apply multi-sample dropout at any of these dropout layers. Multi sampling at an earlier dropout layer may increase diversity among dropout samples and increase the benefits in trade for the higher additional costs due to more duplicated layers. %For example, WideResNet \cite{Zagoruyko16} repeatedly executes dropout since it employs residual blocks which include dropout layers. However, we do not need to duplicate the entire network for applying multi-sample dropout to WideResNet; we can apply the multi sampling at the last residual block for example.

\subsection{Why multi-sample dropout accelerates training}
Intuitively, the effect of multi-sample dropout with $M$ dropout samples is similar to that of enlarging the size of a minibatch $M$ times by duplicating each sample in the minibatch $M$ times, e.g. Batch Augmentation \cite{Hoffer20}. For example, if a minibatch consists of two data samples $\langle A, B\rangle $, training a network by using multi-sample dropout with two dropout samples closely corresponds to training a network by using the original dropout and a minibatch of $\langle A, A, B, B\rangle $ assuming a different mask applied to each sample in the minibatch. This is similar to batch augmentation \cite{Hoffer20}, which applies a different data augmentation for each of duplicated samples to make diversity among duplicated samples. 
Using a larger minibatch size with duplicated samples may not make sense to accelerate the training because it increases the computation time per iteration by $M$ times. In contrast, multi-sample dropout can enjoy similar gains without a huge increase in computation cost per iteration for deep CNNs because it duplicates only the operations after dropout. For example, when we duplicate the last two fully connected layers of VGG16 \cite{Simonyan2014} eight times, we observed the increased execution time per iteration by only 2\%. Because of the non-linearity of the activation functions, the original dropout with duplicated samples and multi-sample dropout do not give exactly the same results. However, similar acceleration was observed in the training in terms of the number of iterations, as shown by the experimental results.

\subsection{Why multi-sample dropout yields higher accuracy}
Noh et al. \cite{Noh17} showed that creating multiple samples during the training of deep networks improves the accuracy of the trained network. Training of a noisy network (e.g. with dropout) requires optimizing the marginal likelihood over the noise ($\mathcal{L}_{marginal}$) and SGD optimizers optimize the network using approximated marginal likelihood based on the finite number of samples ($\mathcal{L}_{SGD}$) as its objective function. Here, the SGD objective function $\mathcal{L}_{SGD}$ is the lower-bound of the marginal likelihood over the noise ($\mathcal{L}_{marginal}$) and using more dropout samples makes the lower-bound tighter, i.e. 

\begin{math}
\mathcal{L}_{marginal} \geq \mathcal{L}_{SGD}(M+1) \geq \mathcal{L}_{SGD}(M),
\end{math}\\here, $\mathcal{L}_{SGD}(M)$ means $\mathcal{L}_{SGD}$ when $M$ dropout samples are used. This results in better accuracy in the trained network with increasing number of dropout samples. Although Noh's Importance Weighted Stochastic Gradient Descent (IWSGD) makes multiple noise (dropout) samples at dropout like our multi-sample dropout, it executes both the forward pass and backward pass separately for each sample, and then calculates the gradients for updating network parameters as weighted average of the gradients calculated for each dropout sample with the normalized likelihood for the sample as the weight. Our results showed that much simpler and light-weight technique which only duplicates a small part of the forward pass can enjoy the benefits of using multiple dropout samples.

% In addition to faster convergence, multi-sample dropout also improves accuracy. Multi-sample dropout inherently creates an ensemble of sub-networks; it generates multiple local loss for dropout samples and then merges them to obtain the final prediction.

\subsection{Other sources of diversity among samples}

The key to faster training with multi-sample dropout is the diversity among dropout samples; if there is no diversity, the multi-sampling technique gives no gain and simply wastes computation resources. Although we tested only dropout in this paper, the multi-sampling technique can be used with other sources of diversity. For example, variants of dropout, such as DropConnect, can be enhanced by using the multi-sampling technique.

% To demonstrate that benefits can be obtained from other sources of diversity, two additional diversity-creation techniques are tested in Appendix: 1) with or without horizontal flipping and 2) zero padding at a pooling layer in left- or right-side of the image. Improvements due to these additional sources of diversity are visible but much less significant than the improvements from dropout. Such results show that dropout is an ideal source of diversity to use in our multi-sampling technique. These additional sources of diversity are not used in the experiments excepts for those in Appendix.
%in the training phase, but not in the inference. %The multi-sampling technique can be also used during inference, but this inference-time ensembling effect is orthogonal to the use of the multi-sampling technique during the training. It is therefore not applied in this paper for fair comparison; we always use only one dropout sample at inference regardless of the training method.
% Random horizontal flipping of the input image is a widely used data augmentation technique in many tasks for image datasets. 
% The multi-sampling technique with flipping and zero padding can be also used during inference to improve the accuracy obtained by ensembling. But this inference-time ensembling effect is orthogonal to the use of the multi-sampling technique during training. It is therefore not applied in this paper for fair comparison; we always use only one dropout sample at inference regardless of the training method.

\section{Experimental Results}

\begin{figure*}
  \centering
  \includegraphics[width=17cm]{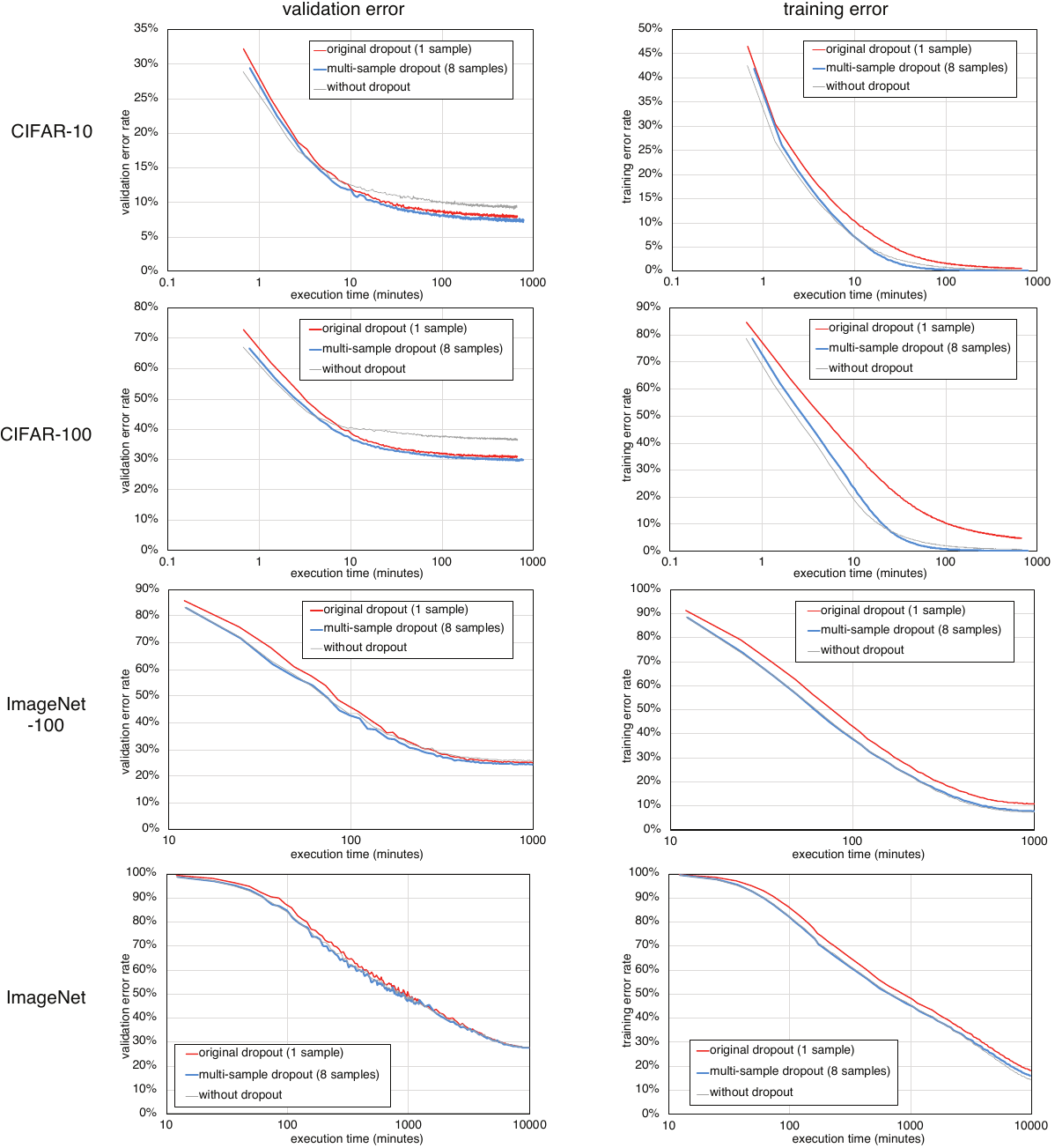}
  \caption{Trends in error rates for validation set and training set against training time for original dropout and multi-sample dropout (in average of five training runs). Multi-sample dropout achieved faster convergence than the original dropout. Note that training errors for multi-sample dropout include the effects of the inherent ensemble mechanism, but no ensemble was used while evaluating validation sets.}
\end{figure*}

\begin{table*}
  \centering
  \caption{Final validation and training error rates of trained networks with original dropout, with multi-sample dropout (with 8 samples) and without dropout (in average of five runs associated with 95\% confidence interval).}
  \includegraphics[width=15cm]{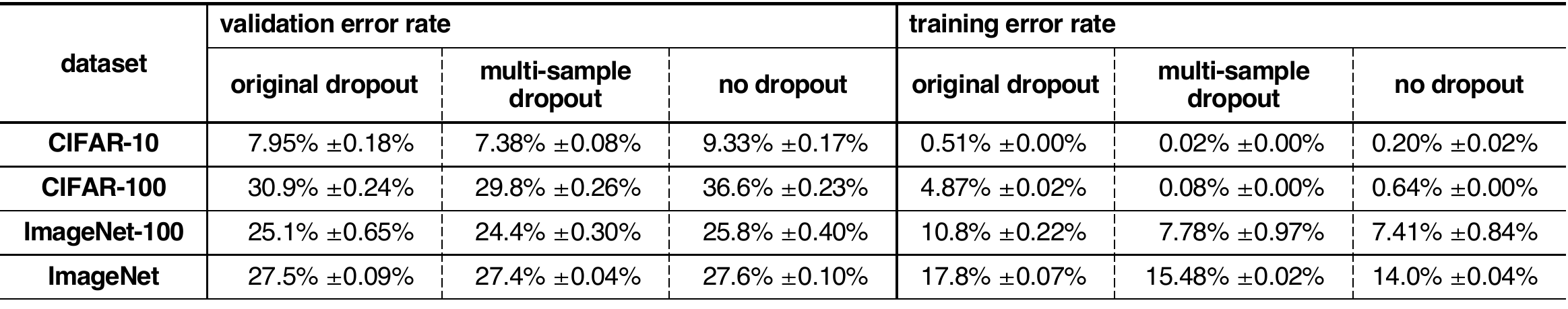}
\end{table*}

\begin{figure*}
  \centering
  \includegraphics[width=14cm]{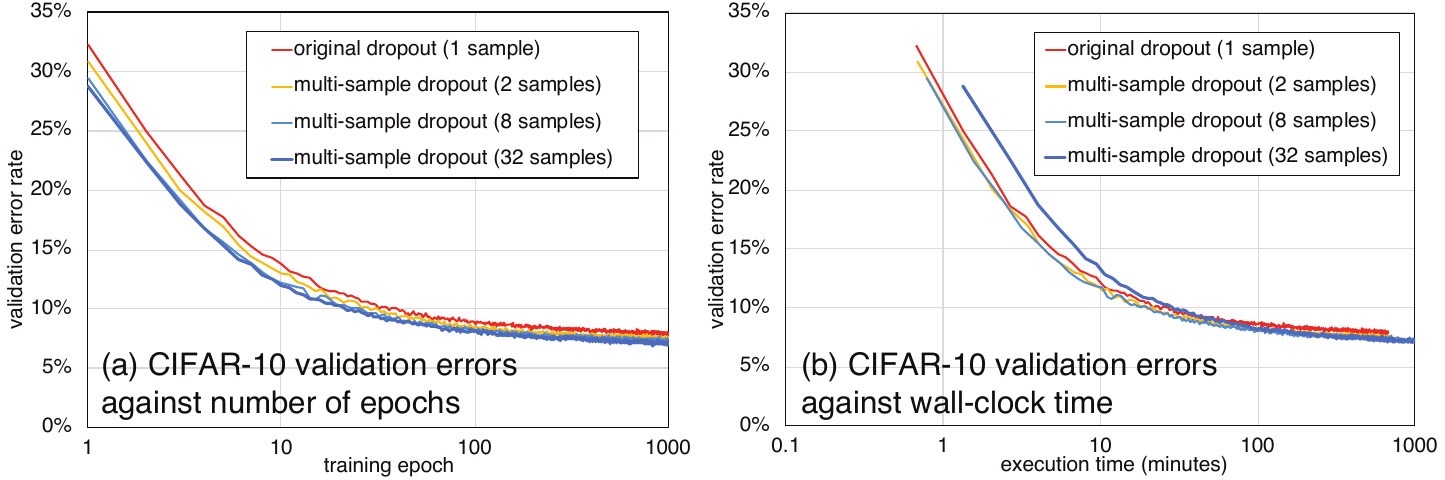}
  \caption{Validation errors during training for CIFAR-10 using different numbers of dropout samples. Using more dropout samples makes convergence faster in terms of number of iterations at the cost of increased execution time per iteration; an excessively large number of dropout samples may hurt the training speed.} %Increasing the number of dropout samples accelerated the training.}
\end{figure*}

\begin{table*}
  \centering
  \caption{Execution time per iteration relative to that of original dropout for different numbers of dropout samples. Increasing the number of dropout samples lengthened the computation time per iteration. We failed to execute VGG16 with 16 dropout samples due to the out-of-memory error with the current minibatch size.}
  \includegraphics[width=15cm]{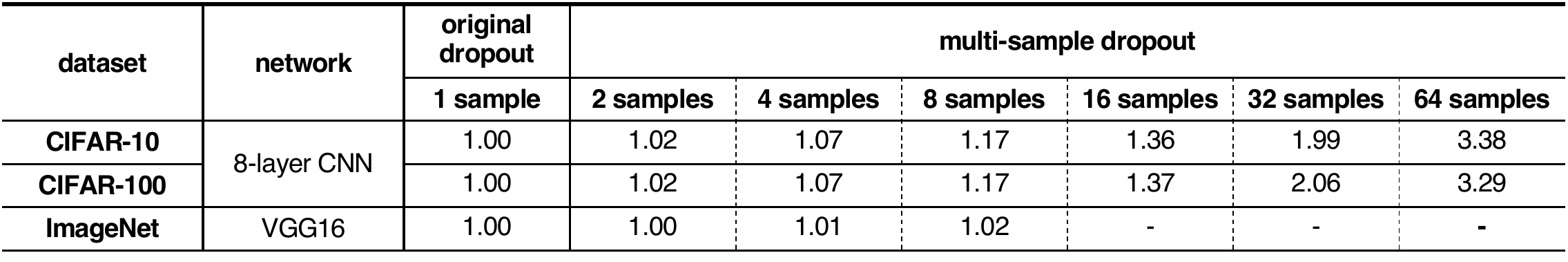}
\end{table*}

\subsection{Implementation}

This section describes the effects of using the multi-sample dropout for various image classification tasks including the ImageNet, CIFAR-10, and CIFAR-100 datasets. For the ImageNet dataset, as well as for the full dataset with 1,000 classes, a reduced dataset with only the first 100 classes was tested (ImageNet-100). For most of the experiments, we use eight as the number of dropout samples, which generally gives good tradeoff between benefits and additional cost.
For the CIFAR-10, and CIFAR-100, an 8-layer network with six convolutional layers and batch normalization \cite{Ioffe15} followed by two fully connected layers with dropout was used. This network executes dropout twice with dropout ratios of 30\%, which are tuned for the original dropout but are used here for all cases unless otherwise specified. The same network architecture except for the number of neurons in the output layer was used for the CIFAR-10 (10 output neurons), and CIFAR-100 (100 output neurons) datasets. The network was trained using the Adam optimizer \cite{Kingma15} with a batch size of 100. These tasks were run on a NVIDIA K20m GPU.% with CUDA 9.0 using the Chainer v4.0 \cite{Tokui2015} as the framework. 
For the ImageNet datasets, VGG16 was used as the network architecture, and the network was trained using stochastic gradient descent with momentum as the optimization method with a batch size of 100 samples. The initial learning rate of 0.01 was exponentially decayed by multiplying it by 0.92 at each epoch. Weight decay regularization was used with a decay rate of $5 \cdot 10^{-4}$ by following the original paper. In the VGG16 architecture, dropout was applied for the first two fully connected layers with 20\% as dropout ratio. A NVIDIA V100 GPU was used for the training with ImageNet datasets. %So far, we have run each experiment only once. with CUDA 10.0 
For all datasets, data augmentation was used by extracting a patch from a random position of the input image and by performing random horizontal flipping during training \cite{Krizhevsky12}. For ImageNet, we additionally apply random resizing and tilting. For the validation set, the patch from the center position was extracted and fed into the classifier without any modifications.
All tests were executed five times and the averages of five results are shown in the figure with 95\% confidence intervals.

% \begin{figure}[t]
%   \centering
% \begin{lstlisting}[basicstyle=\ttfamily\tiny, frame=single]
%   Batch Normalization			28x28x3
%   Convolution 3x3 & ReLU		28x28x64
%   Batch Normalization
%   Convolution 3x3 & ReLU		28x28x96
%   Max Pooling 2x2			14x14x96
%   Batch Normalization
%   Convolution 3x3 & ReLU		14x14x96
%   Batch Normalization
%   Convolution 3x3 & ReLU		14x14x128
%   Max Pooling 2x2			7x7x128
%   Batch Normalization
%   Convolution 3x3 & ReLU		7x7x128
%   Batch Normalization
%   Convolution 3x3 & ReLU		7x7x192
%   for i = 0 to numSamples/2-1:
%     if (i & 1) != 0: HorizontalFlip
%     Max Pooling 2x2			4x4x192
%     Batch Normalization
%     for j = 0 to 1:
%       Dropout 40% dropout ratio
%       if j == 1: HorizontalFlip
%       Fully Connected & ReLU		512
%       Dropout 30% dropout ratio
%       Fully Connected			10 or 100
%       Softmax & CrossEntropy
%   Average all losses
% \end{lstlisting}
% \caption{Design of classification network (with eight weight layers) used for CIFAR and SVHN.}
% \end{figure}

% This section first presents the experimental results to highlight the benefits of multi-sample dropout: accelerated training and better accuracy. It then discusses the effects of the parameters (the number of dropout samples and the dropout ratio) on training speed and accuracy. It also discusses why multi-sample dropout accelerates training.

\subsection{Improvements by multi-sample dropout}

Figure 2 plots the trends in validation errors and training errors against training time for three configurations: trained with the original dropout, with multi-sample dropout, and without dropout. For multi-sample dropout, the losses for eight dropout samples were averaged. How the number of dropout samples affects performance is discussed in the next section. The figure shows that multi-sample dropout achieved faster training than the original dropout for all datasets, i.e. both errors became smaller with the same training time. %The smaller training losses and errors were partially due to its inherent ensemble mechanism.
As is common in regularization techniques, dropout achieves better generalization (i.e., lower validation error rates) compared with the "without dropout" case at the expense of slower training. Multi-sample dropout alleviates this slowdown while still achieving better generalization. 

Table I summarizes the final validation error rates and training error rates. After training, the networks trained with multi-sample dropout were observed to have reduced error rates for all datasets compared with those of the original dropout on average. Note that the improvements in validation errors for ImageNet datasets were not significant based on the confidence intervals.
%The original dropout increased the training errors by avoiding overfitting for all datasets compared with the no dropout case. Multi-sample dropout achieved lower training losses than no dropout for some datasets while avoiding overfitting.

\subsection{Effects of parameters on performance}

\begin{figure*}
  \centering
  \includegraphics[width=13cm]{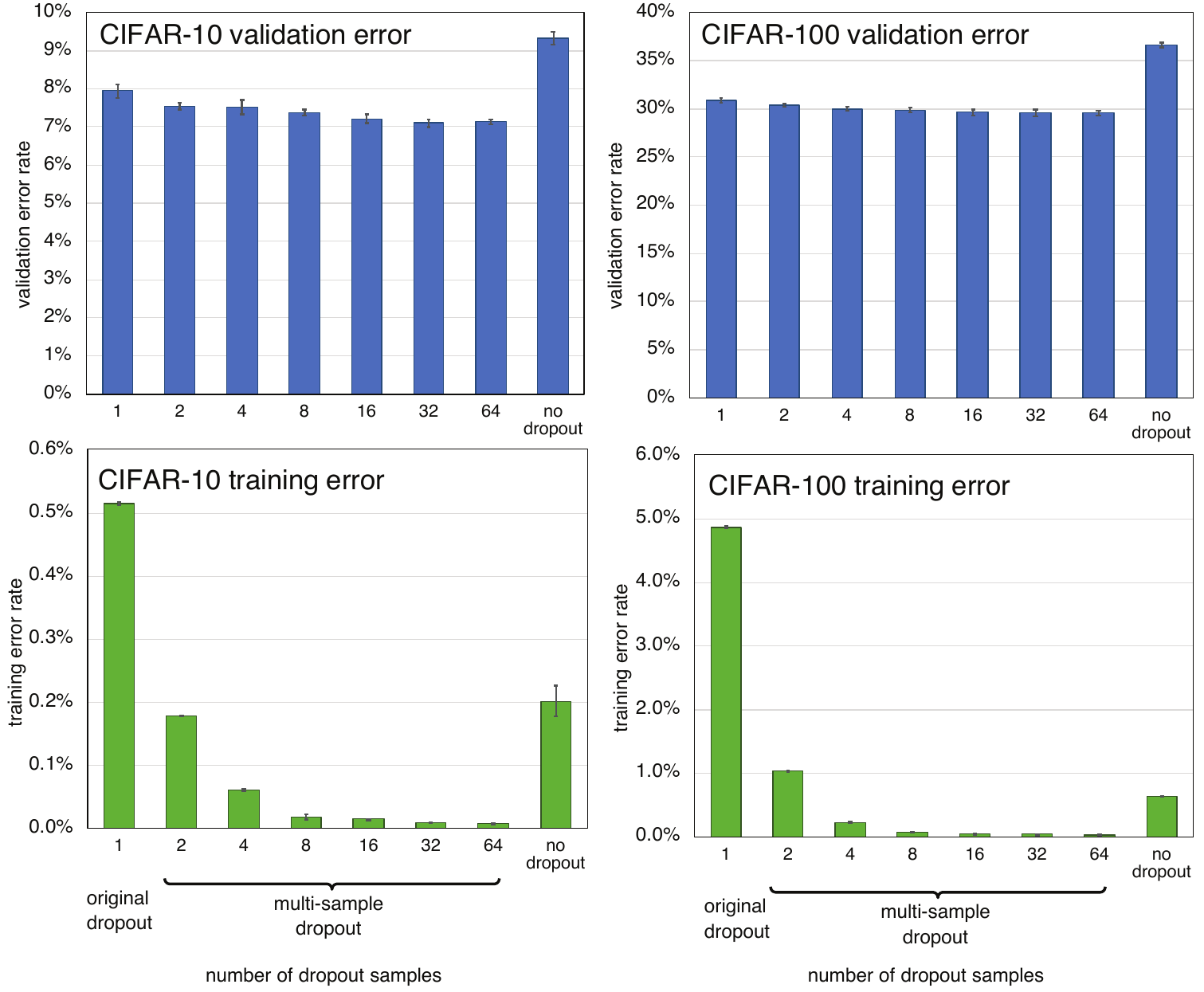}
  \caption{Final validation and training error rates with different numbers of dropout samples. Multi-sample dropout achieved lower error rates with more dropout samples. }
  % \caption{Training losses, training error rates, and validation error rates for different numbers of dropout samples. }
\end{figure*}

% \begin{figure}
%   \centering
%   \includegraphics[width=14cm]{training_speed_time_samples.pdf}
%   \caption{Training losses [and validation errors?] during training for different numbers of dropout samples. Training time was "wall-clock time." }
% \end{figure}

\begin{figure*}
  \centering
  \includegraphics[width=13cm]{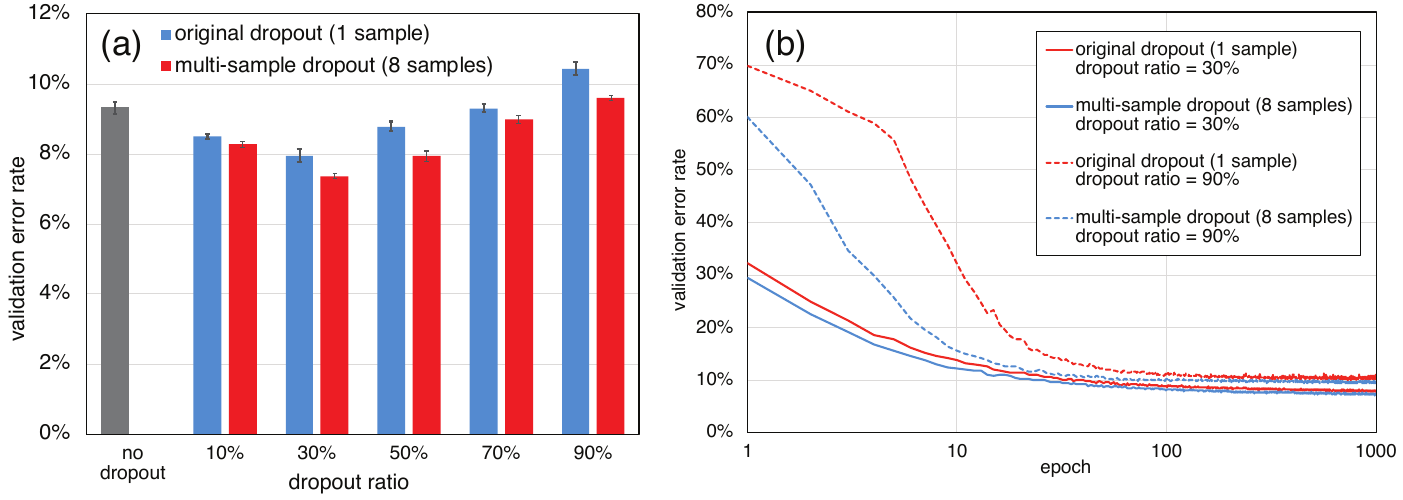}
  \caption{(a) Validation error rates and (b) progress of validation errors during the training with original and multi-sample dropout for various dropout ratios. Multi-sample dropout works regardless of the dropout ratio.}
\end{figure*}

{\bf Number of dropout samples:} %The previous section described the advantages of multi-sample dropout over the original dropout for a configuration with eight dropout samples. This section discusses the effects of the number of dropout samples on performance.
Figure 3(a) and 3(b) compare the validation errors for different numbers of dropout samples (with 1, 2, 8, and 32 samples) for CIFAR-10 against the number of training epochs and wall-clock training time respectively. Using a larger number of dropout samples made training progress faster in terms of the number of epochs (iterations) as well as making the final error rates lower. %With eight dropout samples, for example, the validation accuracy reached 95\% of the final accuracy after the 25th epoch while it was after the 42nd epoch with the original dropout. 
% A clear relationship is evident between the number of dropout samples and the speedup of convergence in terms of the number of epochs (iterations). For the validation error shown in Figure 3(b), the benefit of using more than eight dropout samples was not significant. 

The accelerated training in terms of the number of iterations due to using more dropout samples came at cost: increased execution time per iteration. % and increased memory consumption.
Consideration of the increased execution time per iteration along with the reduced number of iterations revealed that multi-sample dropout achieves the largest speed up in training time when a moderate number of dropout samples, such as 8, is used, as shown in Figure 3(b). Using an excessive number of dropout samples may actually slow down the training.

% Figure 3(c) compare the training error rates for different numbers of dropout samples against the wall-clock time. The trend is similar to that of the validation error rates.

The execution time per iteration relative to that of original dropout is shown in Table II for different numbers of dropout samples. The VGG16 network architecture was used for the ImageNet dataset and a smaller 8-layer CNN was used for the other datasets, as mentioned above. Because a larger network tends to spend more time in deep convolutional layers than in the fully connected layers, which are duplicated in our multi-sample dropout technique, the overhead in execution time compared with that of the original dropout is more significant for the smaller network than it is with the VGG16 network architecture. Multi-sample dropout with eight dropout samples increased the execution time per iteration by about 1.97\% for the VGG16 architecture and by about 17.2\% for the small network. Hence, larger network may benefit from multi-sample dropout more compared to smaller networks. For very smaller networks, the increases in the execution time per iteration may surpass the benefits.% as discussed in Appendix using multilayer perceptron as an example.

The final validation error rates and training error rates of networks trained using different number of dropout samples during the training are shown in Figure 4 for CIFAR-10 and CIFAR-100. %The average losses and error rates between the 1,800th and 2,000th epoch were used.
Multi-sample dropout achieved lower error rates as the number of dropout samples was increased. %The gains were significant when the number was increased from two to four and from four to eight. 
The gains in the validation errors were relatively small when the number was increased above eight considering the increased computation costs shown in Table II. 

From these observations, it was determined that eight is a reasonable value for the number of dropout samples, and it is used in other experiments.

\begin{table*}
  \centering
  \caption{Final validation and training error rates for ResNet with and without multi-sample dropout (using 8 dropout samples).}
  \includegraphics[width=15cm]{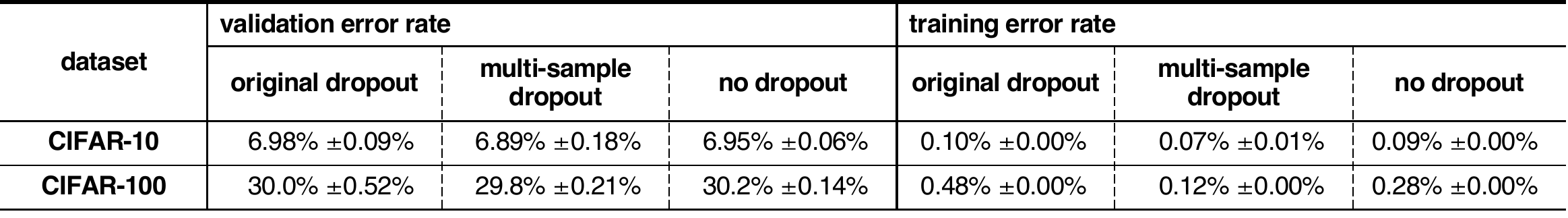}
\end{table*}

\begin{figure*}
  \centering
  \includegraphics[width=14cm]{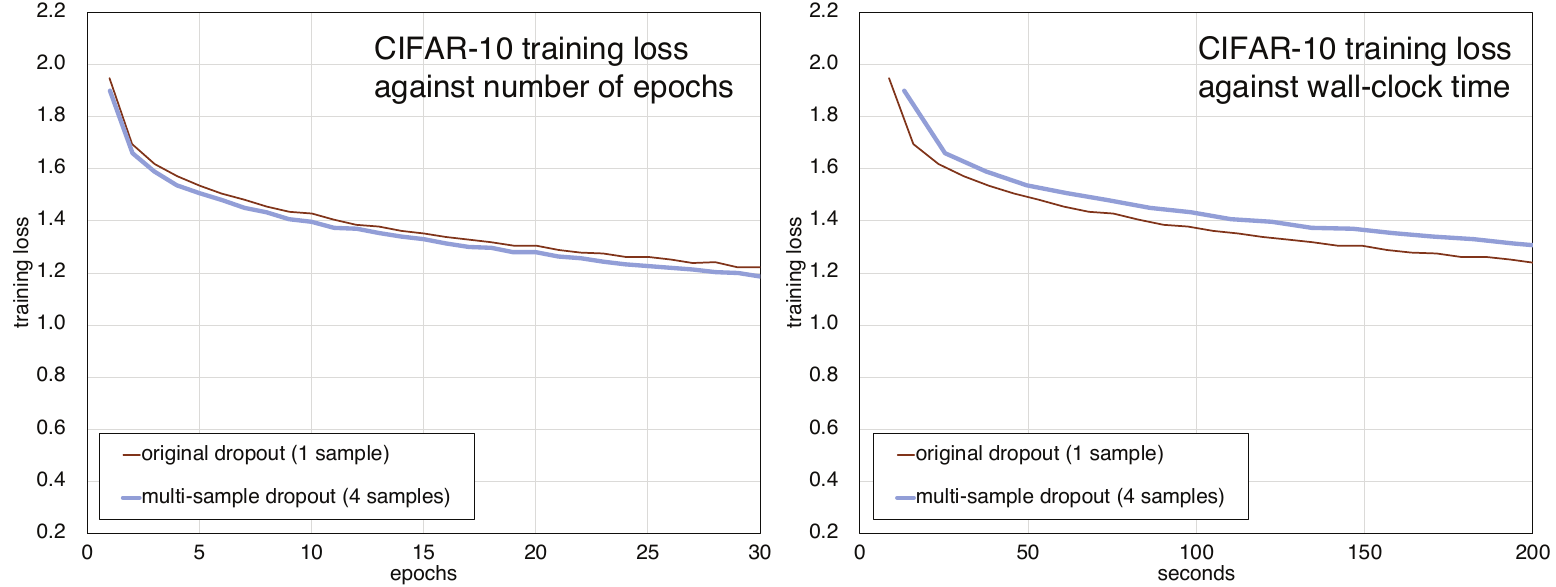}
  \caption{Comparison of training loss using a shallow network (a 4-layer multilayer perceptron) with and without multi-sample dropout for CIFAR-10. The additional computation cost of multi-sample dropout becomes significant in shallow networks while it is quite small for deep convolutional networks. }
\end{figure*}

{\bf Dropout ratio: } Another important parameter is the dropout ratio, which controls the ratio of the neurons to discard. In the 8-layer CNN used for CIFAR datasets, we used 30\% as the ratios in the two dropout layers. This value was tuned for the original dropout but also used for multi-sample dropout. Here we shown how multi-sample dropout works for the CIFAR-10 dataset with various dropout ratios: 10\%, 30\% (default), 50\%, 70\%, and 90\%.

Figure 5(a) shows the final validation error rates. Regardless of the dropout ratio setting, multi-sample dropout consistently achieved lower error rates than the original dropout. When excessively high dropout ratios, such as 90\%, were used, dropout degraded the validation error rate compared with the "without dropout" case. Even with such ratios, multi-sample dropout achieved improvements compared with the original dropout.

Figure 5(b) compares the trend in convergence of validation errors for dropout ratios of 30\% (default) and 90\%. When excessively high dropout ratios of 90\% were used, the speedup by the multi-sample dropout was much more significant.

These results show that multi-sample dropout does not depend on a specific dropout ratio to achieve improvements and that it can be used with a wide range of dropout ratio settings.

\subsection{Effect of multi-sample dropout when original dropout does not work}

Our multi-sample dropout can magnify the benefits of the original dropout. When the original dropout works poorly to improve the accuracy, the multi-sample dropout may also work poorly. For example, it is known that adding a dropout at the end of ResNet architecture \cite{He16}, e.g. after the global average pooling layer, does not improve the final accuracy; hence ResNet typically does not employ dropout after the final pooling layer. We tested ResNet with multi-sample dropout after the pooling layer using CIFAR-10 and CIFAR-100. Table III summarizes the performance with and without multi-sample dropout. The gain from multi-sample dropout in the validation error rates was smaller than the gain with our 8-layer CNN for these datasets (shown in Table I). Whether dropout works well or not depends on many aspects of the workload, e.g. used network architecture, amount of training data, and other regularization techniques (e.g. \cite{Li19}). These characteristics also matters for multi-sample dropout.

\subsection{Applying multi-sample dropout for shallow networks}

As discussed in Section II.B of the paper, multi-sample dropout is mainly targeting deep convolutional neural networks in which most of the computation time is consumed in the convolution layers before the dropout.
Here, we show the effect of multi-sample dropout on shallow networks using a multilayer perceptron as an example. We use a network consists of four fully connected layers each has 2,000 neurons. We apply dropout for each fully connected layer. For multi-sampling, we created four dropout samples at the last fully connected layer; i.e. only one layer is duplicated.
Figure 6 shows the training loss for CIFAR-10 dataset with and without multi-sample dropout. Multi-sample dropout yields smaller training loss compared to the original dropout after the same number of iterations (epochs). However, due to the increase in the computation time per iteration, multi-sample dropout actually degraded the training speed in terms of the training time; the execution time per iteration increased by more than 50\% even we created only four dropout samples. Hence, to make multi-sample dropout effective, it is important to apply multi sampling near the end of the network to limit the number of operations duplicated for multiple dropout samples.

\begin{figure*}
  \centering
  \includegraphics[width=14cm]{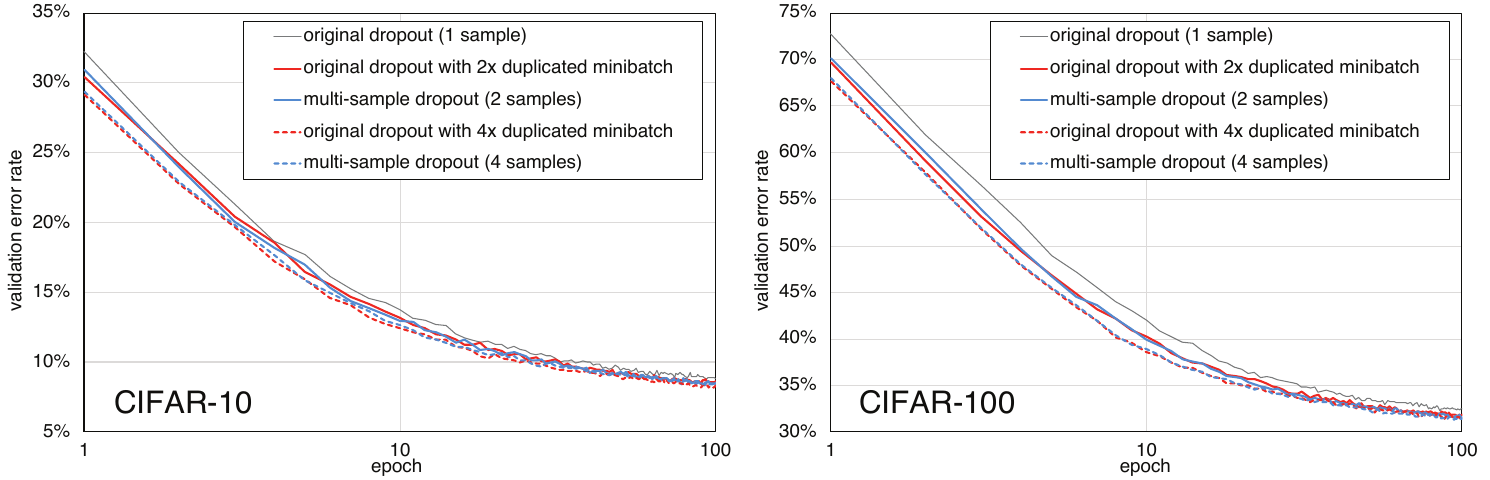}
  \caption{Comparison of original dropout with data duplication in minibatch and multi-sample dropout. X-axis shows number of epochs. Both techniques yield similar improvements in accuracy while the computation cost is much smaller for the multi-sample dropout. }
\end{figure*}

\subsection{Multi-sample dropout and duplicating samples in the same minibatch}

As discussed in Section II.B, the effect of multi-sample dropout with $M$ dropout samples is similar to that of enlarging the size of a minibatch $M$ times by duplicating each sample in the minibatch $M$ times. This is the primary reason for the accelerated training with multi-sample dropout. This is illustrated in Figure 7: the validation errors with multi-sample dropout match well with those of the original dropout using duplicated data in a minibatch. Training errors and training losses also match well although they are not shown here. If the same sample is included in a minibatch multiple times, the results from the multiple samples are ensembled when the parameters are updated, even if there is no explicit ensembling in the network. Duplicating a sample in input and ensembling them at the parameter updates seems to have a quite similar effect on training to that of multi-sample dropout, which duplicates a sample at the dropout and ensembles them at the end of the forward pass. However, duplicating data $M$ times make the execution time per iteration (and hence the total training time) $M$ times longer than without duplication. Multi-sample dropout achieves similar benefits at a much smaller computation cost. %Multi-sample dropout yielded lower training error rates because of its inherent ensemble mechanism at the forward pass.

\section{Related Work}

The multi-sample dropout regularization technique presented in this paper can achieve better generalization and faster training than the original dropout. Dropout is one of the most widely used regularization techniques, but a wide variety of other regularization techniques for better generalization have been reported. They include, for example, weight decay \cite{Krogh91}, data augmentation \cite{Cubuk2018}, \cite{Chawla2002}, \cite{Zhang2017}, \cite{Inoue2018}, label smoothing \cite{Szegedy16}, and batch normalization \cite{Ioffe15}. Although batch normalization is aimed at accelerating training, it also improves generalization. Many of these techniques are network independent while others, such as Shake-Shake \cite{Gastaldi2017} and Drop-Path \cite{Larsson2017}, are specialized for a specific network architecture.

The success of dropout led to the development of many variations that extend the basic idea of dropout (e.g. \cite{Ghiasi18}, \cite{Huang2016}, \cite{Tompson2014}, \cite{DeVries2017}. %Dropout randomly discards half of the neurons at each training iteration to avoid neurons being dependent on each other. 
The techniques reported use a variety of ways to randomly drop information in the network. For example, DropConnect \cite{Wan13} discards randomly selected connections between neurons. DropBlock \cite{Ghiasi18} randomly discards areas in convolution layers while dropout is typically used in fully connected layers after the convolution layers. Stochastic Depth \cite{Huang2016} randomly skip layers in a very deep network. However, none of these techniques use the approach used in our multi-sample dropout. Many of them can be used with multi sampling technique to make the divergence among dropout samples. Another way to enhance the dropout is adaptively tuning the dropout ratio (e.g. \cite{Ba13}. These techniques are also orthogonal to the multi sampling technique, since the multi-sample dropout does not depend on specific dropout ratio as we have already shown.

Multi-sample dropout calculates the final prediction and loss by averaging the results from multiple loss functions. Several network architectures have multiple exits with loss functions. For example, GoogLeNet \cite{Szegedy15} has two early exits in addition to the main exit, and the final prediction is made using a weighted average of the outputs from these three loss functions. Unlike multi-sample dropout, GoogLeNet creates the two additional exits at earlier positions in the network. Multi-sample dropout creates multiple uniform exits, each with a loss function, by duplicating a part of the network.

\section{Summary}

In this paper, we described multi-sample dropout, a regularization technique for accelerating training and improving generalization. The key is creating multiple dropout samples at the dropout layer while the original dropout creates only one sample. Multi-sample dropout can be easily implemented using existing deep learning frameworks by duplicating a part of the network after the dropout layer. Experimental results using image classification tasks demonstrated that multi-sample dropout reduces training time and improves accuracy. Because of its simplicity, the basic idea of the multi-sampling technique can be used in wide range of neural network applications and tasks.

\bibliography{msdo}

\begin{thebibliography}{10}

\bibitem{Ba13}
Lei~Jimmy Ba and Brendan Frey.
\newblock Adaptive dropout for training deep neural networks.
\newblock In {\em Annual Conference on Neural Information Processing Systems
  (NIPS)}, pages 3084--3092, 2013.

\bibitem{Chawla2002}
Nitesh~V. Chawla, Kevin~W. Bowyer, Lawrence~O. Hall, and W.~Philip Kegelmeyer.
\newblock Smote: Synthetic minority over-sampling technique.
\newblock {\em Journal of Artificial Intelligence Research}, 16(1):321--357,
  2002.

\bibitem{Cubuk2018}
Ekin~D. Cubuk, Barret Zoph, Dandelion Mane, Vijay Vasudevan, and Quoc V.~Le.
\newblock Autoaugment: Learning augmentation policies from data.
\newblock {\em arXiv:1805.09501}, 2018.

\bibitem{DeVries2017}
Terrance DeVries and Graham~W. Taylor.
\newblock Improved regularization of convolutional neural networks with cutout.
\newblock {\em arXiv:1708.04552}, 2017.

\bibitem{Gastaldi2017}
Xavier Gastaldi.
\newblock Shake-shake regularization.
\newblock {\em arXiv:1705.07485}, 2017.

\bibitem{Ghiasi18}
Golnaz Ghiasi, Tsung-Yi Lin, and Quoc~V Le.
\newblock Dropblock: A regularization method for convolutional networks.
\newblock In {\em Annual Conference on Neural Information Processing Systems
  (NIPS)}, pages 10727--10737, 2018.

\bibitem{He16}
Kaiming He, Xiangyu Zhang, Shaoqing Ren, and Jian Sun.
\newblock Deep residual learning for image recognition.
\newblock In {\em Conference on Computer Vision and Pattern Recognition
  (CVPR)}, pages 770--778, 2016.

\bibitem{Hinton12}
Geoffrey Hinton, Nitish Srivastava, Alex Krizhevsky, Ilya Sutskever, and Ruslan
  Salakhutdinov.
\newblock Improving neural networks by preventing co-adaptation of feature
  detectors.
\newblock {\em arXiv:1207.0580}, 2012.

\bibitem{Hoffer20}
Elad Hoffer, Tal Ben-Nun, Itay Hubara, Niv Giladi, Torsten Hoefler, and Daniel
  Soudry.
\newblock Augment your batch: Improving generalization through instance
  repetition.
\newblock In {\em Conference on Computer Vision and Pattern Recognition
  (CVPR)}, 2020.

\bibitem{Huang2016}
Gao Huang, Yu~Sun, Zhuang Liu, Daniel Sedra, and Kilian Weinberger.
\newblock Deep networks with stochastic depth.
\newblock {\em arXiv:1603.09382}, 2016.

\bibitem{Inoue2018}
Hiroshi Inoue.
\newblock Data augmentation by pairing samples for images classification.
\newblock {\em arXiv:1801.02929}, 2018.

\bibitem{Ioffe15}
Sergey Ioffe and Christian Szegedy.
\newblock Batch normalization: Accelerating deep network training by reducing
  internal covariate shift.
\newblock {\em arXiv:1502.03167}, 2015.

\bibitem{Kingma15}
Diederik~P. Kingma and Jimmy Ba.
\newblock Adam: A method for stochastic optimization.
\newblock {\em arXiv:1412.6980}, 2014.

\bibitem{Krizhevsky12}
Alex Krizhevsky, Ilya Sutskever, and Geoffrey Hinton.
\newblock Imagenet classification with deep convolutional neural networks.
\newblock In {\em Annual Conference on Neural Information Processing Systems
  (NIPS)}, pages 1106--1114, 2012.

\bibitem{Krogh91}
Anders Krogh and John~A. Hertz.
\newblock A simple weight decay can improve generalization.
\newblock In {\em Annual Conference on Neural Information Processing Systems
  (NIPS)}, pages 950--957, 1991.

\bibitem{Larsson2017}
Gustav Larsson, Michael Maire, and Gregory Shakhnarovich.
\newblock Fractalnet: Ultra-deep neural networks without residuals.
\newblock In {\em International Conference on Learning Representation (ICLR)},
  2017.

\bibitem{Li19}
Xiang Li, Shuo Chen, Xiaolin Hu, and Jian Yang.
\newblock Understanding the disharmony between dropout and batch normalization
  by variance shift.
\newblock In {\em Conference on Computer Vision and Pattern Recognition
  (CVPR)}, 2019.

\bibitem{Noh17}
Hyeonwoo Noh, Tackgeun You, Jonghwan Mun, and Bohyung Han.
\newblock Regularizing deep neural networks by noise: Its interpretation and
  optimization.
\newblock In {\em Annual Conference on Neural Information Processing Systems
  (NIPS)}, pages 5115--5124, 2017.

\bibitem{Simonyan2014}
Karen Simonyan and Andrew Zisserman.
\newblock Very deep convolutional networks for large-scale image recognition.
\newblock {\em arXiv:1409.1556}, 2014.

\bibitem{Szegedy15}
Christian Szegedy, Wei Liu, Yangqing Jia, Pierre Sermanet, Scott Reed, Dragomir
  Anguelov, Dumitru Erhan, Vincent Vanhoucke, and Andrew Rabinovich.
\newblock Going deeper with convolutions.
\newblock In {\em Conference on Computer Vision and Pattern Recognition
  (CVPR)}, 2015.

\bibitem{Szegedy16}
Christian Szegedy, Vincent Vanhoucke, Sergey Ioffe, Jon Shlens, and Zbigniew
  Wojna.
\newblock Rethinking the inception architecture for computer vision.
\newblock In {\em Conference on Computer Vision and Pattern Recognition
  (CVPR)}, 2016.

\bibitem{Tompson2014}
Jonathan Tompson, Ross Goroshin, Arjun Jain, Yann LeCun, and Christopher
  Bregler.
\newblock Efficient object localization using convolutional networks.
\newblock {\em arXiv:1411.4280}, 2014.

\bibitem{Wan13}
Li~Wan, Matthew Zeiler, Sixin Zhang, Yann LeCun, and Rob Fergus.
\newblock Regularization of neural networks using dropconnect.
\newblock In {\em International Conference on Machine Learning (ICML)}, pages
  III--1058--III--1066, 2013.

\bibitem{Zhang2017}
Hongyi Zhang, Moustapha Ciss{\'{e}}, Yann~N. Dauphin, and David Lopez{-}Paz.
\newblock mixup: Beyond empirical risk minimization.
\newblock {\em arXiv:1710.09412}, 2017.

\end{thebibliography}
\bibliographystyle{plain}

\end{document}

\twocolumn[
\icmltitle{Multi-Sample Dropout for Accelerated Training\\ and Better Generalization\\ - Supplementary Material - }



\icmlsetsymbol{equal}{*}

\begin{icmlauthorlist}
\icmlauthor{Hiroshi Inoue}{ibm}
\end{icmlauthorlist}

\icmlaffiliation{ibm}{IBM Research - Tokyo, Tokyo, Japan}

\icmlcorrespondingauthor{Hiroshi Inoue}{inouehrs@jp.ibm.com}

\icmlkeywords{Dropout, Regularization, Convolutional Neural Networks}

\vskip 0.3in
]



\printAffiliationsAndNotice{\icmlEqualContribution} % otherwise use the standard text.

\section{Supplementary Material}

This supplementary material includes some experimental results for backing up the discussion in the main paper.

\section{Applying multi-sample dropout for shallow networks}

\begin{figure*}
  \centering
  \includegraphics[width=14cm]{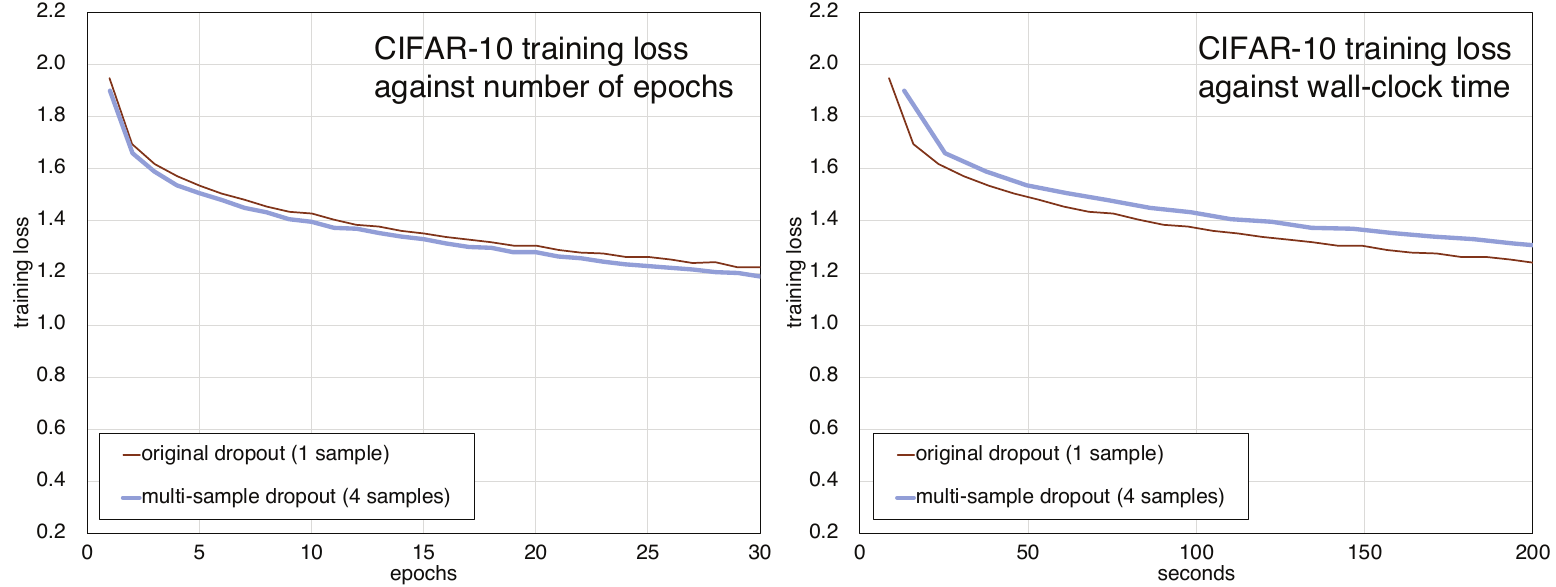}
  \caption{Comparison of training loss using the multilayer perceptron with and without multi-sample dropout for CIFAR-10. }
\end{figure*}

As discussed in Section 2.1 of the paper, multi-sample dropout is mainly targeting deep convolutional neural networks in which most of the computation time is consumed in the convolution layers before the dropout.
Here, we show the effect of multi-sample dropout on shallow networks using a multilayer perceptron as an example. We use a network consists of four fully connected layers each has 2,000 neurons. We apply dropout for each fully connected layer. For multi-sampling, we created four dropout samples at the last fully connected layer; i.e. only one layer is duplicated.
Figure 1 shows the training loss for CIFAR-10 dataset with and without multi-sample dropout. Multi-sample dropout yields smaller training loss compared to the original dropout after the same number of iterations (epochs). However, due to the increase in the computation time per iteration, multi-sample dropout actually degraded the training speed in terms of the training time; the execution time per iteration increased by more than 50\% even we created only four dropout samples. Hence, to make multi-sample dropout effective, it is important to apply multi sampling near the end of the network to limit the number of operations duplicated for multiple dropout samples.

\section{Improvements from other sources of diversity among dropout samples}

As discussed in Section 2.3, zero padding at a pooling layer and horizontal flipping can be used as sources of diversity among dropout samples in addition to the different masks in dropout. 

In the current implementation, the horizontal flipping is applied immediately before the first fully connected layer; a half of the dropout samples are horizontal flipped deterministically. When pooling an image with a size that is not a multiple of the window size, e.g., when applying 2x2 pooling to a 7x7 image, the zero padding can be added on the left or right and at the top or bottom. Here, zero padding is added on the right (and bottom) for half of the dropout samples and on the left (and bottom) for the other half. By doing these operations in addition to using a different mask for each dropout sample, the diversity among dropout samples are further enhanced.
%The location of the zero padding is controlled by using horizontal flipping; i.e., "flip, pool with zero padding on right, and then flip" is equivalent to "pool with zero padding on left." %This paper also describes how these two types of modifications can be integrated.
% If the trained network is used for inference as is, it inherently executes ensemble of multiple samples with and without flipping (and padding on left and on right). 

Figure 2 shows how these additional sources contribute to training acceleration. Although their contributions to the training speed are not negligible, they are much smaller than those of dropout. Nevertheless, these results show that the multi-sampling technique can work with not only dropout but also with other sources of divergence among samples while the dropout is an ideal source of divergence among multiple samples. 

\begin{figure*}
  \centering
  \includegraphics[width=13cm]{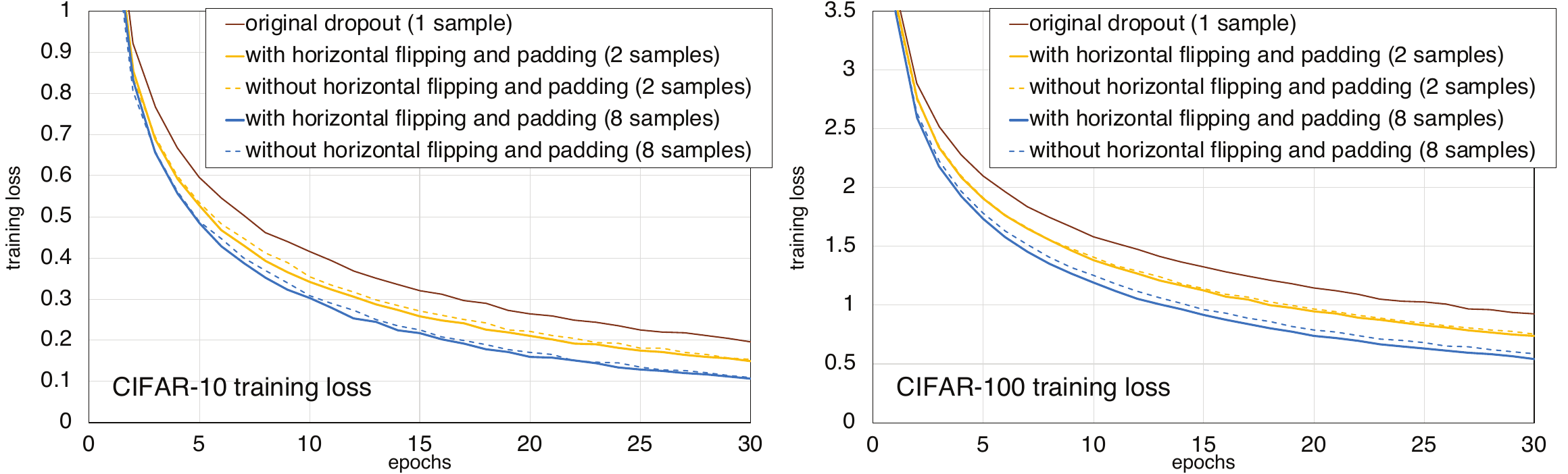}
  \caption{Comparison of training loss with and without horizontal flipping to create additional diversity among dropout samples in multi-sample dropout. X-axis shows number of epochs. }
\end{figure*}